\documentclass[akbc,twoside,11pt,lettersize]{article}
\usepackage{akbc}
\akbcheading
\ShortHeadings{Knowledge Base Question Answering: A Semantic Parsing Perspective}{Gu, Pahuja, Cheng \& Su}
\finalcopy 

\usepackage{times}
\usepackage{xspace}
\usepackage{multirow}
\usepackage{wrapfig}
\usepackage[table,xcdraw,dvipsnames]{xcolor}
\newcommand{\nop}[1]{}
\newenvironment{remark}[1][Remark]{\begin{trivlist}
\item[\hskip \labelsep {\bfseries #1}]}{\end{trivlist}}

\newenvironment{entityfont}{\fontfamily{qcr}\selectfont\footnotesize}{\par}
\newenvironment{relationfont}{\fontfamily{qcr}\selectfont\footnotesize}{\par}
\DeclareTextFontCommand{\textentity}{\entityfont}
\DeclareTextFontCommand{\textrelation}{\relationfont}

\newcommand{\Freebase}{\textsc{Freebase}\xspace}
\newcommand{\Wiki}{\textsc{Wikidata}\xspace}
\newcommand{\DBpedia}{\textsc{DBpedia}\xspace}

\newcommand{\WebQ}{\textsc{WebQ}\xspace}
\newcommand{\WebQSP}{\textsc{WebQSP}\xspace}
\newcommand{\ComplexQ}{\textsc{ComplexWebQ}\xspace}
\newcommand{\FreeCai}{\textsc{Freebase917}\xspace}
\newcommand{\GraphQ}{\textsc{GraphQ}\xspace}
\newcommand{\Grail}{\textsc{GrailQA}\xspace}

\newcommand{\QALD}{\textsc{QALD}\xspace}
\newcommand{\CSQA}{\textsc{CSQA}\xspace}
\newcommand{\KQA}{\textsc{KQA Pro}\xspace}
\newcommand{\LCQ}{\textsc{LC-QuAD}\xspace}

\newcommand{\Picard}{\textsc{Picard}\xspace}
\newcommand{\BigBird}{\textsc{BigBird}\xspace}
\newcommand{\Grappa}{\textsc{Grappa}\xspace}

\include{todonotes}

\usepackage{soul}
\usepackage{url}
\usepackage{hyperref}
\usepackage[utf8]{inputenc}
\usepackage[small]{caption}
\usepackage{graphicx}
\usepackage{amsmath}
\usepackage{amsthm}
\usepackage{booktabs}
\usepackage{algorithm}
\usepackage{algorithmic}
\begin{document}

\title{Knowledge Base Question Answering: A Semantic Parsing Perspective}

\author{\name Yu Gu \email gu.826@osu.edu \\
       \addr Department of Computer Science and Engineering\\
       The Ohio State University, USA
       \AND
       \name Vardaan Pahuja \email pahuja.9@osu.edu\\
       \addr Department of Computer Science and Engineering\\
       The Ohio State University, USA
       \AND
       \name Gong Cheng \email gcheng@nju.edu.cn\\
       \addr State Key Laboratory for Novel Software Technology\\
       Nanjing University, China
       \AND
       \name Yu Su \email su.809@osu.edu \\
       \addr Department of Computer Science and Engineering\\
       The Ohio State University, USA}



\maketitle

\begin{abstract}
Recent advances in deep learning have greatly propelled the research on semantic parsing. Improvement has since been made in many downstream tasks, including natural language interface to web APIs, text-to-SQL generation, among others. However, despite the close connection shared with these tasks, research on question answering over knowledge bases (KBQA) has comparatively been progressing slowly. We identify and attribute this to two unique challenges of KBQA, \textit{schema-level complexity} and \textit{fact-level complexity}. In this survey, we situate KBQA in the broader literature of semantic parsing and give a comprehensive account of how existing KBQA approaches attempt to address the unique challenges. Regardless of the unique challenges, we argue that we can still take much inspiration from the literature of semantic parsing, which has been overlooked by existing research on KBQA. Based on our discussion, we can better understand the bottleneck of current KBQA research and shed light on promising directions for KBQA to keep up with the literature of semantic parsing, particularly in the era of pre-trained language models.
\end{abstract}

\section{Introduction}
\begin{figure}[!t]
    \centering
    \includegraphics[width=0.64\columnwidth]{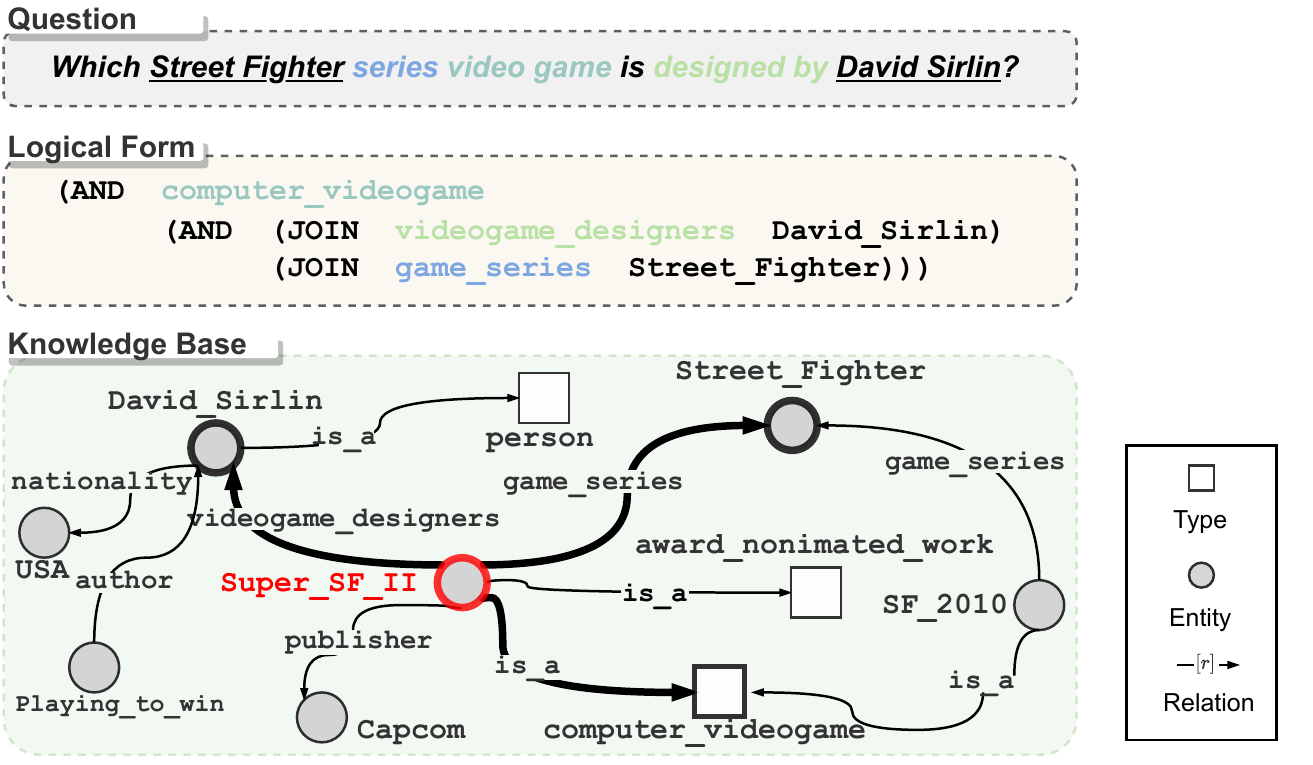}
    \caption{Most state-of-the-art approaches to KBQA are based on semantic parsing, i.e., a question is translated into a logical form (an S-expression~\protect\cite{gu2020beyond} in this example), which is then executed over the KB to retrieve the answer (i.e., \textentity{Super\_SF\_II}). The semantic parsing task (implicitly) entails learning an alignment between the utterance and the KB schema (i.e., \emph{schema linking}). We highlight the alignment using the same color. Named entities (marked with underlines) are usually linked beforehand to reduce the search space. Finally, the logical form composed from the linked items should be able to be grounded to the KB (i.e., \emph{faithfulness}), where we highlight the grounding with bold lines.}
    \label{fig:example}
\end{figure}

Modern knowledge bases (KBs) like \Freebase~\cite{bollacker2008freebase} and \Wiki~\cite{vrandevcic2014wikidata} store abundant world facts and organize them structurally following a crafted schema (i.e., an ontology). Knowledge base question answering (KBQA), which aims to locate the answers from the KB given a question in natural language, provides users with easy access to such massive structured data in KBs by offering a unified natural language interface to shield users from the heterogeneity underneath. State-of-the-art\nop{\gong{I'm not sure whether IR-based methods can be better. Actually, we don't have to claim that semantic parsing represents SOTA.}} approaches to KBQA are predominantly based on semantic parsing,\nop{\footnote{In this survey, we abuse the term semantic parsing to refer to executable semantic parsing~\cite{liang2016learning}\nop{, where a backend database is available for execution}.}} i.e., a question is mapped onto a logical form (e.g., SPARQL~\cite{ravishankar2021two} or $\lambda$-Calculus~\cite{cai2013semantic}) that can be executed against the backend (i.e., a KB) to retrieve the answers (see Figure~\ref{fig:example}).

Neural models~\cite{dong-lapata-2016-language,jia-liang-2016-data,hwang2019comprehensive,wang-etal-2020-rat} have greatly enhanced the performance of semantic parsing, benefiting from recent advances such as the encoder-decoder framework~\cite{seq2seq,bahdanau2014neural} and pre-trained language models (PLMs)~\cite{devlin-etal-2019-bert,t5}. However, these advances have mainly inspired semantic parsing over a backend with a relatively simple schema (e.g., relational databases~\cite{yu-etal-2018-spider} and web APIs~\cite{su2017building}). They have been largely under-exploited\nop{\ysu{I'd be cautious about the word ``overlooked''. There are plenty of neural KBQA methods. It's more like under-exploited.}} in KBQA due to several unique challenges facing large-scale KBs\nop{, which hinder KBQA from readily exploiting\nop{\ysu{exploiting}} them}. 

Semantic parsing in KBQA is uniquely challenging for two main reasons: \emph{schema-level complexity} and \emph{fact-level complexity}. On the one hand, the schema of a modern KB is extremely rich. For example, \Freebase has over 8K schema items in total (6K relations and 2K types), while a relational database usually comprises dozens of\nop{\ysu{there could be more (e.g., hundreds) in a real DB. It's just that the text-to-SQL datasets have less. don't be too absolute.}} schema items only (i.e., table names and column headers)~\cite{yu-etal-2018-spider}. Therefore, learning an alignment between natural language and the schema (i.e., schema linking) is much more challenging in KBQA. Also, the KB ontology models more complex relationships among schema items, (e.g., type hierarchy and domain/range information of relations). Therefore, a deeper understanding of the schema is in demand for KBQA compared with other semantic parsing tasks.\nop{; and current techniques like joint encoding for schema items simply via concatenation using PLMs~\cite{hwang2019comprehensive}, which is commonly used in text-to-SQL, would fall short of this goal. }

\nop{\gong{If space allows, I would like to see a figure illustrating the two challenges with toy examples.}}

On the other hand, although content-agnostic models suffice to produce good results in other semantic parsing tasks, fact-level information (i.e., contents in the database) plays an integral role in KBQA. This is because the schema of a KB is instantiated dynamically. For instance, in \Freebase's schema, the type \textrelation{person} can be associated with over 1K different relations, while each instance of \textrelation{person} may only be associated with a few of them (e.g., \textrelation{David\_Sirlin} in Figure~\ref{fig:example} only has 4 relations). Consequently, generating logical forms that can ground to non-empty answers from the KB, i.e., \textit{faithful} to the KB, requires tightly incorporating fact-level information. In addition, the graph structure of KB facts leads to an enormous search space due to combinatorial explosion, rendering generating faithful queries even more challenging. As a result, the encoder-decoder framework, which has been the \textit{de facto} choice for many state-of-the-art semantic parsers, can be hardly adopted for KBQA.

\nop{
\paragraph{Existing Solutions.}
\ysu{Existing Solutions? Also, is it necessary to discuss these here? What value does it add?}
A potpourri of approaches have been proposed to address the uniquely challenging task of KBQA. We categorize them into three families: \emph{ranking methods}, \emph{coarse-to-fine methods}, and \emph{generation methods}. In spite of their disparate designs, these methods share the same overarching idea, i.e., they all take the structure of the KB (either schema-level or fact-level) as a prior\ysu{not clear what ``prior'' here means exactly} for semantic parsing, rather than delegating the semantic parsing task to a purely data-driven learning model\ysu{this is vague. all ML are data-driven learning.}. Specifically, ranking methods\nop{~\cite{yih-etal-2015-semantic,berant-liang-2014-semantic,hu-etal-2018-state,lan2019knowledge,gu2020beyond}} first enumerate a set of candidate queries directly from the KB and then semantic parsing boils down to computing the matching score for each candidate-question pair. Coarse-to-fine methods\nop{~\cite{ding2019leveraging,sun2020sparqa,hua2020less,hu2021edg}} first generate possible query skeletons and then ground the skeletons to the KB with admissible schema items. Generation methods\nop{~\cite{liang-etal-2017-neural,chen-etal-2019-uhop,lan-jiang-2020-query}}, which emerge more recently, typically ground a query to the KB on the fly\ysu{on the fly, otherwise it's an adjective}\gong{I think the term 'generation' has explained itself, while on the contrary the explanation of 'on-the-fly' makes me confused.}, and thus dynamically reduce the search space. These methods all leverage KB structure to reduce the search space for schema linking, hence handling the schema-level complexity. Also, they address the fact-level complexity and produce faithful queries using pre-processing (i.e., candidate enumeration), post-processing (i.e., skeleton grounding), and online processing respectively.
}

\begin{figure}[!t]
    \centering
    \includegraphics[width=0.4\columnwidth]{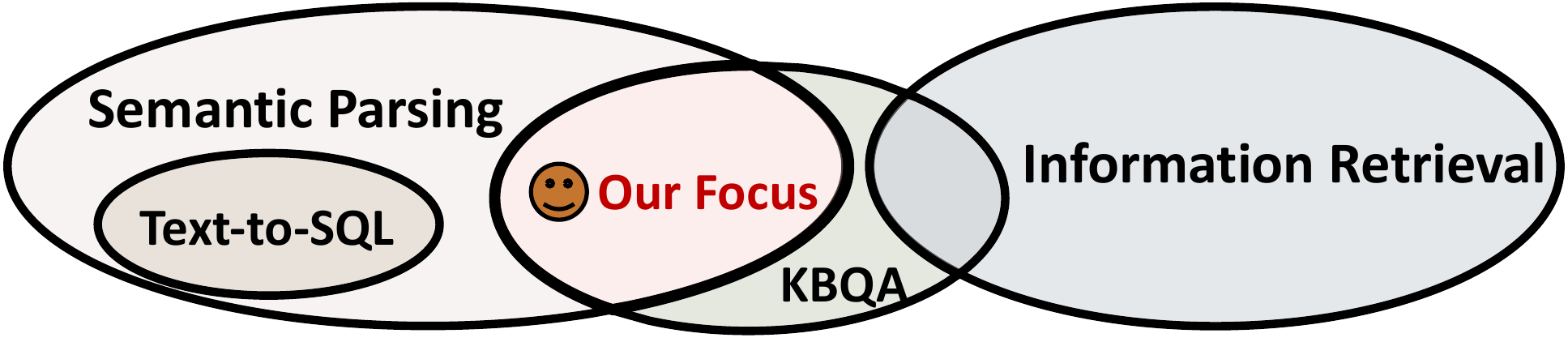}
    \caption{We survey KBQA research based on semantic parsing and draw insights from neighboring tasks (e.g., text-to-SQL) in the broad literature of semantic parsing.}
    \label{fig:venn}
\end{figure}

\nop{\gong{I would strongly recommend adding a figure here to illustrate the positioning of this paper. It could be a Venn figure showing relationships between terms: KBQA, semantic parsing, text-to-SQL, IR-based methods, etc. \response{ysu}{that's a good idea}}}
We discuss the recent progress in KBQA from a semantic parsing perspective. Different from existing surveys which either endeavor to offer an inclusive overview of works on KBQA from different communities~\cite{diefenbach2018core,chakraborty2021introduction}, or take a task-oriented view and focus on a specific branch of KBQA (i.e., KBQA with complex questions~\cite{fu2020survey,lansurvey,lan2021survey}), we focus on semantic parsing-based KBQA, the most prevailing KBQA methods, and discuss both challenges and solutions with our proposed taxonomy---therefore a technique-oriented view---while other methods (e.g., information retrieval based KBQA) are not our focus (see Figure~\ref{fig:venn}). Notably, previous surveys only discuss existing works within the KBQA realm, while no effort has been made to situate KBQA in the broader literature of semantic parsing. Our survey fills this gap, which is critical for understanding the limits of current KBQA research and identifying future opportunities by taking inspiration from recent trends in semantic parsing. \nop{Figure~\ref{fig:venn} illustrates the positioning of this survey.}

 
\section{Background}
\label{sec:background}
\subsection{Knowledge Base}
A knowledge base (KB) $\mathcal{K}$ stores facts in a structured way. It comprises two parts: an ontology $\mathcal{O}$ and a model $\mathcal{M}$. The ontology $\mathcal{O}$ defines rules about how facts should be organized in $\mathcal{K}$ (i.e., it defines the class hierarchy and domain/range information for relations). $\mathcal{M}$ instantiates $\mathcal{O}$ and represents facts, where each fact is a triple of the form \textit{(subject, relation, object)}. Formally, $\mathcal{O}\subseteq\mathcal{C}\times\mathcal{R}\times\mathcal{C}$ and $\mathcal{M} \subseteq \mathcal{E}\times\mathcal{R}\times(\mathcal{C}\cup\mathcal{E}\cup\mathcal{L})$, where $\mathcal{E}$ is a set of entities, $\mathcal{R}$ is a set of binary relations, $\mathcal{C}$ is a set of classes, and $\mathcal{L}$ is a set of literals, and $\mathcal{C}\cup\mathcal{R}$ constitute the schema items of $\mathcal{K}$. Compared with relational databases, KBs feature much more sophisticated schemas. Also, facts are typically organized into a graph rather than rows in a table. KBs thus encode more structured information.

\subsection{KBQA Task}
\begin{remark}[Task formulation.]
Given $\mathcal{K}=\mathcal{O}\cup\mathcal{M}$ and a question $q$ in natural language, typically, the task of KBQA is to find a set of entities $\mathcal{A}\subseteq\mathcal{E}$, with $|\mathcal{A}|\geq 1$,\nop{\gong{Aggregation queries should be explicitly excluded here.}} that capture the intent of $q$. $\mathcal{A}$
can be directly returned as the answer or further aggregated (e.g., applying a counting function). For example, given a question \textit{``Which street Fighter series video game is designed by David Sirlin?''} the goal is to find the answer entity \textrelation{Super\_SF\_II}, with $\mathcal{K}$ being  \Freebase (see Figure~\ref{fig:example}). 
\end{remark}

\begin{remark}[Logical forms.] 
Logical forms (i.e., queries) for KBQA can be in different meaning representations (MRs), e.g., SPARQL~\cite{yih-etal-2016-value}, graph query~\cite{su2016generating}, $\lambda$-DCS~\cite{cai2013semantic}, and S-expression~\cite{gu2020beyond}.
\nop{Different MRs are mutually-convertible.\ysu{too absolute. perfect convertibility is rare.}} $\lambda$-DCS and S-expression can be viewed as the linearized version of graph query and better suits encoder-decoder models. Also, they can be viewed as the syntactic sugar for SPARQL.\footnote{This claim does not hold generally because the different expressiveness of MRs, but holds in our context because KBQA usually only exploit a subset of expressiveness for most MRs.} An example of a query in S-expression is depicted in Figure~\ref{fig:example}.\nop{ and the graph query counterpart is in Figure~\ref{fig:frameworks}. \ysu{make sure you update Figure 2 or remove this}} A \emph{well-formed} query is syntactically correct and thus can be executed with no exception, but possibly with an empty answer, while a \emph{faithful} query must find non-empty answers from the KB.

\end{remark}

\begin{remark}[Evaluation.]
To benchmark the research on KBQA, a line of datasets have been released during the past decade (summarized in Table~\ref{table:datasets}). Questions in all datasets except for \CSQA, which only contains machine-generated questions, are either collected from web search logs (i.e., \WebQ) or paraphrased by crowd workers. Most datasets are typically collected with synthetic queries with pre-defined templates or random sampling. Different datasets may serve different purposes. For example, \ComplexQ is a dataset that particularly focuses on complex questions. \Grail is a dataset that can systematically evaluate the generalizability of KBQA models beyond i.i.d. level. 

Generally, there are two types of metrics for evaluation: execution-based metrics (e.g., $\boldsymbol F_1$), which encourage partially correct prediction, and more strict metrics based on logical form (e.g., \textbf{ExactMatch}).\nop{ \ysu{the datasets need a bit more elaboration.}}

\nop{Most datasets typically first generate a set of queries via pre-defined templates~\cite{talmor2018web,dubey2018earl} or randomly sampling over the KB~\cite{su2016generating,gu2020beyond} and then invite crowd workers to annotate natural language questions (usually via paraphrasing). Most datasets hold an i.i.d. assumption.
Gold queries are available in most datasets except for \WebQ and \CSQA, where \WebQ contains question-answer pairs collected from the web search logs and thus features the most realistic distribution. A subset of \WebQ are annotated with gold queries by experts and form \WebQSP. Other}


\begin{table}[!t]
    \small
    \centering
    \resizebox{.85\textwidth}{!}{
    \begin{tabular}{c|cccc}
    \toprule
        \textbf{Dataset} & \textbf{KB} &\textbf{Size} &
        \textbf{Logical Form} & \textbf{Generalization Assumption}\\ \midrule
        \QALD~\cite{ngomo20189th} & \DBpedia& 558& SPARQL & comp. \\
         \LCQ~\cite{trivedi2017lc} & \DBpedia & 5,000&SPARQL & i.i.d. \\
        \LCQ 2.0~\cite{dubey2019lc}& \Wiki, \DBpedia & 30,000& SPARQL & i.i.d. \\
        \CSQA~\cite{saha2018complex} & \Wiki & 800,000 & N/A & i.i.d.\\
        \KQA~\cite{shi2020kqa} & \Wiki &117,970 &Program & i.i.d.\\
        \FreeCai~\cite{cai2013large} & \Freebase & 917& $\lambda$-Calculus & zero-shot \\
        \WebQ~\cite{berant-etal-2013-semantic} & \Freebase & 5,810 & N/A & i.i.d. \\
        \WebQSP~\cite{yih-etal-2016-value} & \Freebase &4,737 & SPARQL & i.i.d. \\
        \ComplexQ~\cite{talmor2018web} & \Freebase& 34,689& SPARQL& i.i.d.\\
        \GraphQ~\cite{su2016generating} & \Freebase& 5,166 & Graph query & comp.+zero-shot \\
        \Grail~\cite{gu2020beyond} & \Freebase& 64,331& S-expression & i.i.d.+comp.+zero-shot \\
    \bottomrule
    \end{tabular}}
    \caption{Information on existing KBQA datasets. For the generalization assumption, we follow the definitions by~\protect\citet{gu2020beyond} \nop{\ysu{probably need to be double-column. Explain what ``-'' mean.}}}
    \label{table:datasets}
\end{table}
\end{remark}


\section{Approaches}
\label{sec:methods}



\begin{table*}[!t]
    \small
    \centering
    \resizebox{\textwidth}{!}{
    \begin{tabular}{ccc}
    \toprule
    \multicolumn{1}{c}{\textbf{Categories}} & \multicolumn{1}{c}{\textbf{Sub-tasks}} & \textbf{Solutions} \\ 
                    \midrule
        \multicolumn{1}{c|}{\multirow{6}{*}{Ranking}}&  \multicolumn{1}{c|}{\multirow{3}{*}{\emph{Candidate Enumeration}}}& \multirow{2}{45em}{Template-based: \citet{bast2015more}, \citet{berant-liang-2014-semantic}, \citet{abujabal2017automated}; KB traversal: \citet{yih-etal-2015-semantic}, \citet{zafar2018formal}, \citet{hu-etal-2018-state}, \citet{lan2019knowledge}, \citet{gu2020beyond}$^\heartsuit$, \citet{ye2021rng}$^\heartsuit$}\\
        \multicolumn{1}{c|}{}& \multicolumn{1}{c|}{} & \\
        \multicolumn{1}{c|}{}& \multicolumn{1}{c|}{} & \\
        \cmidrule{2-3}
        \multicolumn{1}{c|}{}&  \multicolumn{1}{c|}{\multirow{3}{*}{\emph{Semantic Matching}}}& \multirow{2}{45em}{Learning-to-rank: \citet{bast2015more}, \citet{abujabal2017automated}, \citet{yih-etal-2015-semantic}, \citet{zafar2018formal}, \citet{hu-etal-2018-state}, \citet{lan2019knowledge}; PLMs: \citet{gu2020beyond}$^\heartsuit$, \citet{ye2021rng}$^\heartsuit$; Paraphrasing: \citet{berant-liang-2014-semantic}}\\
        \multicolumn{1}{c|}{}& \multicolumn{1}{c|}{} & \\
        \multicolumn{1}{c|}{}& \multicolumn{1}{c|}{} & \\
        \midrule

         \multicolumn{1}{c|}{\multirow{6}{*}{Coarse-to-fine}}&  \multicolumn{1}{c|}{\multirow{3}{*}{\emph{Skeleton Parsing}}}& \multicolumn{1}{c}{\multirow{3}{45em}{Question decomposition: \citet{ding2019leveraging}, \citet{bhutani2019learning}, \citet{hu2021edg}$^\heartsuit$; Encoder-decoder: \citet{ravishankar2021two}$^\heartsuit$, \citet{das-etal-2021-case}$^\heartsuit$, \citet{zhang2019complex}; Pipeline:\citet{sun2020sparqa}$^\heartsuit$; AMR parsing: \citet{kapanipathi-etal-2021-leveraging}, \citet{bornea2021learning}$^\heartsuit$}}\\
         \multicolumn{1}{c|}{}& \multicolumn{1}{c|}{} & \\
         \multicolumn{1}{c|}{}& \multicolumn{1}{c|}{} & \\
        \cmidrule{2-3}
        \multicolumn{1}{c|}{}&  \multicolumn{1}{c|}{\multirow{3}{*}{\emph{Grounding}}}& \multirow{2}{45em}{Filling partial results: \citet{ding2019leveraging}, \citet{bhutani2019learning}, \citet{hu2021edg}$^\heartsuit$; Refining: \citet{ravishankar2021two}$^\heartsuit$, \citet{das-etal-2021-case}$^\heartsuit$; Using training data: \citet{sun2020sparqa}$^\heartsuit$ Tranpiling: \citet{kapanipathi-etal-2021-leveraging}, \citet{bornea2021learning}$^\heartsuit$}; \\
        \multicolumn{1}{c|}{}& \multicolumn{1}{c|}{} & \\
        \multicolumn{1}{c|}{}& \multicolumn{1}{c|}{} & \\
        \midrule

        \multicolumn{1}{c|}{\multirow{4}{*}{Generation}}& \multicolumn{1}{c|}{\multirow{4}{*}{-}}
        & \multirow{4}{45em}{Graph search: \citet{lan2019multi}, \citet{chen-etal-2019-uhop}, \citet{lan-jiang-2020-query}$^\heartsuit$; Unconstrained decoding: \citet{zhang2019complex}, \citet{yin2021510},
        \citet{gu2020beyond}$^\heartsuit$, \citet{modern}$^\heartsuit$; Schema-level constrained decoding: \nop{\citet{hua2020less}, }\citet{chen-etal-2021-retrack}$^\heartsuit$, \citet{cao2021program}$^\heartsuit$; Fact-level constrained decoding: \citet{liang-etal-2017-neural}, \citet{ansari2019neural}, \citet{qiu2020hierarchical},
        \citet{gu2022arcaneqa}$^\heartsuit$}
  \\
        \multicolumn{1}{c|}{}& \multicolumn{1}{c|}{} & \\
        \multicolumn{1}{c|}{}& \multicolumn{1}{c|}{} & \\
        \multicolumn{1}{c|}{}& \multicolumn{1}{c|}{} & \\

    \bottomrule
    \end{tabular}}
    \caption{We summarize representative KBQA studies based on semantic parsing. Existing works fall into three families: \emph{ranking}, \emph{coarse-to-fine}, and \emph{generation} methods. Methods using PLMs are indicated with $^\heartsuit$.}
    \label{table:approaches}
\end{table*}

\begin{wrapfigure}{L}{0.5\textwidth}
    \centering
    \includegraphics[width=0.5\columnwidth]{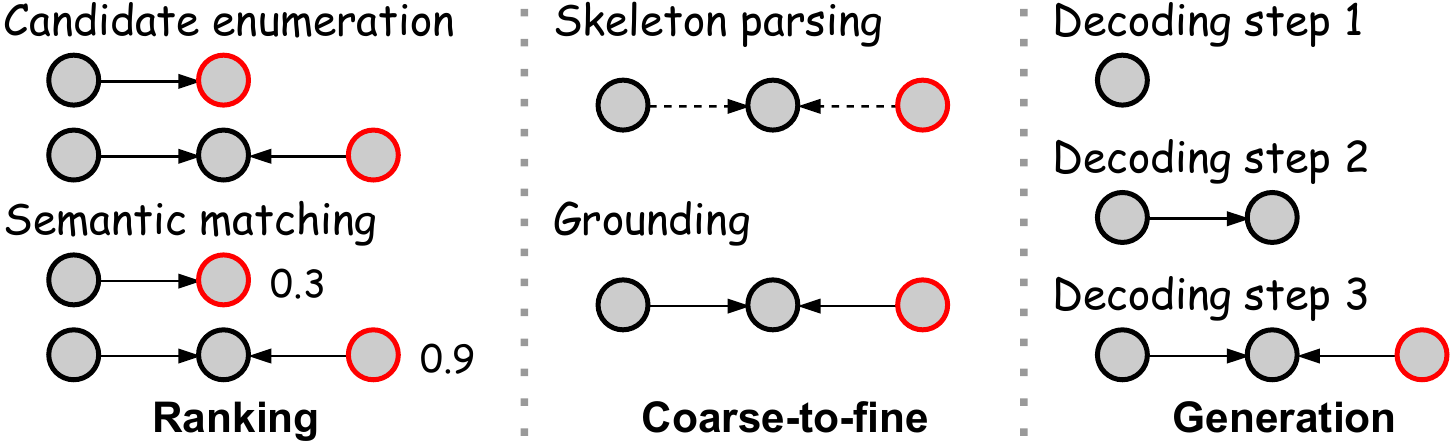}
    \caption{Illustrations of the high-level ideas of three semantic parsing-based KBQA families. Each query is represented as a graph (here a path for brevity). Entities or query nodes are both represented as a circle, where the red circle denotes the target node. An edge denotes a KB relation while a dashed edge denotes a relation placeholder to be grounded.}
    \label{fig:frameworks}
\end{wrapfigure}

A potpourri of approaches have been proposed to address the uniquely challenging task of KBQA. We categorize them into three families: \emph{ranking methods}, \emph{coarse-to-fine methods}, and \emph{generation methods} (Figure~\ref{fig:frameworks}). Despite the disparate designs, all methods share the same overarching idea, i.e., using the KB structure (either schema-level or fact-level) to constrain output space.  Specifically, ranking methods first enumerate candidate queries from the KB and semantic parsing thus boils down to computing the matching score for each candidate-question pair. Coarse-to-fine methods first generate query skeletons and then ground the skeletons to the KB with admissible schema items. Generation methods, which have emerged more recently, typically ground a query to the KB via constrained decoding, and thus dynamically reduce the search space. These methods all leverage the KB structure to reduce the search space for schema linking, hence handling the \textit{schema-level complexity}. Also, they address the \textit{fact-level complexity} and produce faithful queries using pre-processing (i.e., candidate enumeration), post-processing (i.e., skeleton grounding), and online processing respectively (i.e., constrained decoding). Tabel~\ref{table:approaches} summarizes existing works in different families.


\subsection{Ranking Methods}
Ranking methods decompose KBQA into two disjointed sub-tasks, i.e., \emph{candidate enumeration} and \emph{semantic matching}. Candidate enumeration incorporates fact-level information by directly listing faithful queries from the KB. Semantic matching aims at returning a matching score for each question-candidate pair and is usually modeled as a machine learning task.
\paragraph{Candidate Enumeration.}
Traditional methods pre-specify a set of query templates and then fill the templates with concrete arguments. For example, \citet{bast2015more} and \citet{berant-liang-2014-semantic} only define three and five query templates respectively and fill the template slots for entities or schema items using heuristics\nop{ over surface forms}. \citet{abujabal2017automated} mine query templates from training data and do slot filling with a lexicon. Another common strategy is to first identify a topic entity from the question and then use it as an anchor to traverse its neighborhood in the KB to enumerate possible queries. For example, \citet{yih-etal-2015-semantic} \nop{and \citet{luo2018knowledge}} first enumerate core inferential chains (i.e., relation paths) starting from the topic entity and then extend the chains by adding possible constraints or aggregation functions to form candidate queries. \citet{hu-etal-2018-state} extend their work to enumerate more diverse candidate queries using crafted operations over the KB (e.g., merge and expand)\nop{, namely, connect, merge, expand, and fold}. Similarly, \citet{gu2020beyond} and \citet{ye2021rng} also enumerate relation paths starting from identified entities\nop{ and then apply aggregations (e.g., \textentity{COUNT} in \Grail) to the relation paths}. \citet{lan2019knowledge} seek to use more anchors other than named entities, such as entities of common nouns and relation types, to guide the candidate enumeration, while \citet{zafar2018formal} also use a set of candidate relations to generate candidate queries. Though these methods can effectively use identified anchors for faithful query enumeration, they all suffer from scalability, e.g., they typically impose a maximum length limit of 2 for relation paths.
\paragraph{Semantic Matching.}
Early methods rely on hand-crafted features and learning-to-rank approaches to find the top-ranked candidate~\cite{bast2015more,yih-etal-2015-semantic,abujabal2017automated,hu-etal-2018-state,zafar2018formal,lan2019knowledge}. Commonly used features include simple statistics such as the number of grounded entities and features retrieved by neural models, e.g., \citet{yih-etal-2015-semantic} use a CNN model to compute a matching score for each pair of question and core relation, while a Tree-LSTM~\cite{tai-etal-2015-improved} is used by~\citet{zafar2018formal}. Recent methods start to use PLMs for semantic matching. \citet{gu2020beyond} first train a BERT-based Seq2Seq model and then use it as a scorer, while \citet{ye2021rng} directly use BERT to get the matching score by linearizing each candidate query and formulating each question-candidate pair as a sentence pair input to BERT. \citet{berant-liang-2014-semantic} pursue a different direction by converting candidate queries into canonical utterances and modeling the semantic matching task as paraphrasing.

\subsection{Coarse-to-fine Methods}
Coarse-to-fine methods decompose semantic parsing into two stages. First, a model only predicts a rough query skeleton (or a sketch), which focuses on the high-level structure (\emph{a coarse parse}). Second, the model fills the missing details by grounding the query skeleton to the KB (\emph{a fine parse}). The intuition is to disentangle information at different levels of granularity in semantic parsing~\cite{dong-lapata-2018-coarse,zhang2019complex}. In addition, the two-stage searching more effectively prunes the search space resulting from combinatorial explosion in KBQA, compared with ranking methods.
\paragraph{Skeleton Parsing.}
\nop{Some works merge a query skeleton from substructures and potentially entail a question decomposition task. For instance,} \citet{ding2019leveraging} learn frequent query substructures from training data and generate a query skeleton that merges them. \citet{hu2021edg} rely on manually-crafted rules operating over constituency parses to generate a special query skeleton, named entity description graph \nop{(i.e., entities with attributes described in natural language)}. \citet{bhutani2019learning} predict a computation plan for the question, which indicates how a complex question should be decomposed into several sub-questions. Different from them, \citet{sun2020sparqa} perform skeleton parsing in a pipeline with sub-tasks like question split and span prediction. Recent works predict a skeleton from the question using encoder-decoders. \citet{ravishankar2021two} decode a sketch for SPARQL with textualized relation placeholders that can support cross-KB generalization. \citet{bornea2021learning,kapanipathi-etal-2021-leveraging} delegate skeleton parsing to an AMR parser pre-trained on external corpora. \nop{Different from all these methods, }\citet{das-etal-2021-case} use T5~\cite{t5} to directly output a coarse query from the question and retrieved cases from training data. The query may contain inaccurate schema items that need further revision during grounding. 


\paragraph{Grounding.}
The grounding step fills (or revises) a query skeleton to produce the final faithful query. \citet{ding2019leveraging} use off-the-shelf systems to link possible entities and relations in the question, and then try all possible combinations of them for each skeleton. The final score is based on both the linking probability and skeleton parsing probability. \citet{hu2021edg} ground the skeleton by mapping entity attributes in natural language into KB relations using BERT as a binary classifier. \citet{bhutani2019learning} ground their computation plan by predicting each partial query using a semantic matching module based on LSTM and word embeddings. \citet{sun2020sparqa} find the most similar question from training data and use the schema items in that question to ground the skeleton. \citet{ravishankar2021two} rely on a BERT-based encoder to map the KB-agnostic relation placeholders onto a concrete relation in the target KB. \citet{bornea2021learning,kapanipathi-etal-2021-leveraging} transpile AMR into SPARQL queries with specified mapping rules and neural modules for relation linking and decoding. \citet{das-etal-2021-case} revise the inaccurate schema items by aligning them with items in the neighborhood of the topic entity using both string-level and embedding-level similarities.


\subsection{Generation Methods}\label{subsec:gen}
Generation methods have been the de facto choice for many semantic parsing tasks and have also been a trending paradigm for KBQA due to their flexibility and scalability. Adapting generation methods to KBQA poses a unique challenge for producing faithful queries, which requires tightly incorporating the KB structure during generation (decoding). 
\paragraph{Graph Search Paradigm.}
Several works capitalize on the intuition that a faithful query can be produced with direct graph search over the KB given anchor entities\nop{\cite{lan2019multi,chen-etal-2019-uhop,lan-jiang-2020-query}}. \citet{lan2019multi,chen-etal-2019-uhop} propose to perform beam search over the KB to find the top-$K$ relation paths, i.e., at each step, they rank the reachable relations given the context and extend the beam paths with the top-ranked ones. \citet{lan-jiang-2020-query} extend this line of work by introducing more operations for graph search. In addition to extending the paths, they define two more actions: \emph{connect} and \emph{aggregate}, where connect adds an extra constraint to a partial query and aggregate assigns an aggregation function over variables. A key question for these methods is how to condition the search process on the input question. Also, the conditioning should be dynamic during the search process. \citet{chen-etal-2019-uhop} propose a dynamic question representation module which generates a new question representation at each search step. \citet{lan-jiang-2020-query} use BERT as a sentence-pair classifier to provide a matching score feature between each partial query and the question at each step.
\paragraph{Encoder-Decoder Paradigm.} 
Encoder-decoders have actually offered standard solutions not only to conditioning the search on the input question via attention mechanism~\cite{bahdanau2014neural,vaswani2017attention} but also to modeling decoding history and termination. \citet{yin2021510} directly applies an encoder-decoder framework to translate a question into a SPARQL query, where SPARQL's syntactic symbols are preprocessed to ease the learning. \citet{modern} applies more advanced pre-trained encoder-decoders like T5 and BART~\cite{lewis-etal-2020-bart}, assuming entities and relations in the target query are given as input. However, both methods adopt an unconstrained decoder which predicts free-formed queries from the enormous search space with no faithfulness guarantee\nop{ (e.g., \citet{zhang2019complex}), which is the key challenge for using encoder-decoders in KBQA}. \citet{gu2020beyond} propose to build a question-specific decoding vocabulary, where only schema items reachable from the identified entities within 2 hops are included. Their method can significantly reduce the search space but has guarantees for neither well-formedness nor faithfulness. \citet{chen-etal-2021-retrack} and \citet{cao2021program} impose schema-level constraints during decoding to prune the search space with a grammar-based decoder and a function-based decoder respectively. Queries predicted by them are well-formed but may still be unfaithful. To provide a faithfulness guarantee, a more effective solution is to impose fact-level constraints for prediction. Specifically, \citet{liang-etal-2017-neural} propose to predict a query token by token, where a set of admissible tokens\footnote{A token is admissible if it can lead to a faithful query. Their entire token vocabulary comprises schema items, intermediate variables, and syntactic items like function names or brackets.} is derived based on the decoding history and intermediate executions at each step. \citet{ansari2019neural,qiu2020hierarchical} share the same spirit but instead of operating over the tokens space, they predict an action (e.g., choosing a function or an argument) at each step to better suit semantic parsing. \citet{gu2022arcaneqa} also adopt token-based constrained decoding, while they propose a novel contextualized encoder fueled by PLMs and their decoder can support more types of KB queries.
\nop{\citet{hua2020less} similarly propose to either predict an action or copy an item from the candidate schema items recognized during a pre-processing step, however, their pipeline design is bound to error propagation. }

\subsection{Training}
KBQA models are trained with either \emph{strong supervision} (i.e., question-query pairs) or \emph{weak supervision} (i.e., question-answer pairs). The choice of supervision can be orthogonal to the model architecture. Training from weak supervision demands effectively searching for a set of proxy target queries, after which the optimization process resembles learning with strong supervision, with the difference being that each proxy query is weighed based on the similarity between its execution and the gold answers~\cite{yih-etal-2015-semantic,liang-etal-2017-neural,lan-jiang-2020-query}. Searching for proxy queries can be modeled with reinforcement learning, where the weights define the rewards. One major challenge arises from the sparse reward signal at the early stages. To address it, different techniques, such as pre-training~\cite{qiu2020hierarchical} and iterative ML training~\cite{liang-etal-2017-neural}, have been proposed to warm up the model\nop{ to search for better proxy queries}. Note that which type of supervision is more advantageous remains to be further investigated. \citet{yih-etal-2016-value} point out that training with weak supervision may yield sub-optimal results due to spurious proxy queries, while \citet{wang2022new} find that using multiple proxy queries provides more comprehensive information and leads to better performance. \nop{What supervision is more advantageous on existing benchmarks remains to be further investigated.}

\subsection{Empirical Results}
\begin{wraptable}{R}{0.45\textwidth}
    \small
    \centering
    \resizebox{.45\textwidth}{!}{
    \begin{tabular}{c|cc}
    \toprule
        \textbf{Dataset} & \textbf{Top-$\boldsymbol 1$} $\boldsymbol F_1$ & \textbf{Top-$\boldsymbol 1$ Family}\\ \midrule
         \LCQ & 75.0\cite{zafar2018formal} & Ranking \\
        $\KQA^\clubsuit$ & 90.6\cite{lewis-etal-2020-bart}$^\heartsuit$ & Generation\\\
        \WebQSP & 76.5\cite{cao2021program}$^\heartsuit$ & Generation \\
        $\ComplexQ^\clubsuit$&70.0\cite{das-etal-2021-case}$^\heartsuit$ & Coarse-to-fine\\
        \GraphQ & 31.8\cite{gu2022arcaneqa}$^\heartsuit$ & Generation \\
        $\Grail^\clubsuit$ &  74.4\cite{ye2021rng}$^\heartsuit$ & Ranking\\
    \bottomrule
    \end{tabular}}
    \caption{We present the best-published results on KBQA benchmarks with at least two semantic parsing-based models evaluated on them. $^\clubsuit$ indicates benchmarks with official evaluation scripts. $^\heartsuit$ indicates using PLMs.}
    \label{table:empirical}
\end{wraptable}
\nop{We also present empirical results of recent works on several popular KBQA benchmarks to provide a better understanding of the capacity of different families of KBQA models.} In Table~\ref{table:empirical}, we show the $\boldsymbol F_1$ and the corresponding family of best-performing models on KBQA benchmarks that have at least two published semantic parsing-based models evaluated. On \KQA, \WebQSP, and \GraphQ, the state-of-the-art models are based on generation~\cite{lewis-etal-2020-bart,cao2021program,gu2022arcaneqa}, while ranking methods achieve the best results on \LCQ and \Grail~\cite{zafar2018formal,ye2021rng} and the best performance on \ComplexQ is obtained by the coarse-to-fine method~\cite{das-etal-2021-case}. Though no family dominates all the benchmarks, generation methods tend to be a trending option due to the easy integration of PLMs---all best-performing models are based on PLMs except for \LCQ.\footnote{There is no published result on \LCQ using PLMs so far.} We also want to note that the results on \KQA are relatively higher and the results on \GraphQ are relatively lower because \KQA uses a down-sampled KB which leads to smaller search space while \GraphQ only evaluates on challenging non-i.i.d. questions with a small training set of 2583 training questions. For more empirical results, we refer readers to~\citet{perevalov2022knowledge}.



\section{Semantic Parsing Literature}\label{sec:literature}

We briefly review research on semantic parsing, focusing on several recent trends. By drawing insights from trends in the broader literature of semantic parsing, we can better understand the bottleneck of current KBQA research and discuss promising future directions (Section~\ref{sec:discussion}).

\paragraph{From Pipeline to End-to-End.}
Traditional semantic parsing methods are typically based on pipelines. For example, early works first build a lexicon of phrases paired with meaning representations. Then, given an input utterance, they identify relevant lexical entries from the lexicon and apply combination rules to synthesize the logical form in a bottom-up manner \cite{zettlemoyer2005uai,cai2013semantic,berant-etal-2013-semantic}. Both constructing a lexicon and specifying combination rules require domain knowledge and thus suffer from flexibility, e.g., \citet{berant-etal-2013-semantic} use domain-specific corpora like ClueWeb to construct their lexicon, while \citet{zettlemoyer2005uai} and \citet{cai2013semantic} need to define a \nop{combinatory categorial grammar}CCG for combination rules. To provide a more general solution, \citet{dong-lapata-2016-language} adapt the Seq2Seq model to address semantic parsing by modeling it as a sequence transduction task. The end-to-end encoder-decoder paradigm has since been the de facto choice for many semantic parsing tasks. However, a vanilla Seq2Seq model which predicts free-formed sequences can be sub-optimal for semantic parsing. To better adapt to the encoder-decoder paradigm for semantic parsing, a common strategy is to adopt a grammar-based decoder which only outputs well-formed queries~\cite{krishnamurthy2017neural,wang-etal-2020-rat}. More recently, \citet{rubin-berant-2021-smbop} propose a novel semi-autoregressive bottom-up decoder that achieves better efficiency with parallelization.

\paragraph{Semantic Parsing with Pre-Training.}
PLMs are versatile for a wide range of NLP tasks due to their general knowledge of language. Unsurprisingly, they have also received much success in semantic parsing. \nop{For example, on the leaderboard of Spider,\footnote{https://yale-lily.github.io/spider} almost all top-ranked models are based on PLMs.} Encoder-decoder-based semantic parsers usually use PLMs to provide better contextualized representations for both the question and the schema of the target backend. Specifically, a common practice is to concatenate the question and all schema items together as the input to the PLM, and thus contextualization is achieved via the PLM's self-attention layers~\cite{hwang2019comprehensive}. Recent works have also used PLMs for decoding. Encoder-decoder PLMs are trained with unstructured textual data and have an unconstrained output space, which is different from semantic parsing. To deal with it, \Picard\cite{scholak2021picard} finds well-formed output for text-to-SQL using PLMs by rejecting inadmissible tokens at each decoding step. Specifically, two different levels of checking are applied. First, lexical-level checking helps to reject invalid tokens from the output space (e.g., a misspelled column header). Second, schema-level checking helps to reject things like selecting a column from the table to which it does not belong. 

In addition, continuing pre-training PLMs with an in-domain corpus (i.e., task-specific pre-training) can better tailor PLMs for semantic parsing~\cite{yu2020grappa,herzig-etal-2020-tapas,deng-etal-2021-structure,liu2021tapex}. For example, \Grappa\cite{yu2020grappa} synthesizes an in-domain corpus of utterance-SQL pairs generated from a synchronous context-free grammar and proposes a binary classification task for pre-training called SQL semantic prediction (SSP), which predicts whether a table column appears in the target SQL query.  With the task-specific corpus and objective, \Grappa considerably improves the performance on text-to-SQL over general-purpose PLMs.

\paragraph{Out-of-Distribution Neural Semantic Parsing.}
Different from traditional semantic parsing methods~\cite{zettlemoyer2005uai,cai2013semantic}, neural semantic parsers usually hold an i.i.d. assumption~\cite{dong-lapata-2016-language,hwang2019comprehensive}. Semantic parsers that operate with an i.i.d. assumption may fail in real-life scenarios where true user distribution is hard to capture. In addition, training semantic parsers with weak generalizability can be data-inefficient. To support the study of out-of-distribution generalization for semantic parsing, several benchmarks have been released. Particularly, \citet{yu-etal-2018-spider} release Spider to evaluate cross-domain generalization in text-to-SQL. Each question in Spider is given its own target database, and databases in the test set are never seen during training. In this setting, models are encouraged to really perform semantic parsing rather than pattern memorization, i.e., the target query for a question may be seen during training under i.i.d. setting, which offers semantic parsers a shortcut to make predictions.

\section{Discussion}\label{sec:discussion}
\nop{PLM\\
Beam search based on encoder decoder\\
Task-specific Pre-training\\
Interaction \\
Joint entity linking and semantic parsing \\
Hybrid QA \\
differentiate reasoning (william cohen)\\
strong generalizability \\
Linked open data (beyond kb generalization)}

\nop{\gong{While the discussions are helpful, one may ask why these issues are selected instead of others. Ideally, these issues should be selected such that they are closely related to the conclusions drawn from the previous section. \response{agreed. overall there's lack of connections/echos between different sections so it doesn't quite form a cohesive story yet.}}}

To conclude this survey, we provide in-depth discussions on promising directions in KBQA, tightly drawing insights from the semantic parsing research discussed in Section~\ref{sec:literature}. \nop{We hope our survey can give readers a better understanding of not only the KBQA research but also the semantic parsing literature. }

\subsection{Towards End-to-End KBQA} \label{subsec:e2e}
\paragraph{KBQA Based on Encoder-Decoders}
Despite the fact that the encoder-decoder paradigm has revolutionized the research in semantic parsing and become a norm for many downstream tasks, it is not as popular in KBQA. This is due to the challenge in generating faithful queries for KBQA using encoder-decoder models, as we discussed earlier in Section~\ref{subsec:gen}. Here we elaborate more on the difficulty in generating faithful queries for KBQA and text-to-SQL using encoder-decoder models. Specifically, encoder-decoder models can make predictions with three different levels of control, namely, unconstrained decoding, decoding with schema-level constraints, and decoding with fact-level constraints. For text-to-SQL, unconstrained decoding can achieve satisfactory results under the i.i.d. setting~\cite{hwang2019comprehensive}, while for KBQA, a Seq2Seq model with unconstrained decoding considerably underperforms the ranking methods~\cite{gu2020beyond}. This is because KBs feature a much more sophisticated schema whose structures cannot be well learned in a data-driven manner using an unconstrained decoder. We can inject prior knowledge on the schema to constrain the decoding output space. For example, in text-to-SQL such constraints include selecting a column from the table it belongs to, while in KBQA such constraints can be defined based on a relation's domain/range information. Such schema-level constraints guarantee the well-formedness of both text-to-SQL and KBQA and can be easily implemented with grammar-based decoders~\cite{krishnamurthy2017neural}. For existing text-to-SQL benchmarks, well-formedness almost equals to faithfulness, while this does not hold for KBQA because a KB is instantiated dynamically, e.g., not every \textrelation{Person} is associated with the relation \textrelation{videogame\_designers}. As a result, more complicated fact-level constraints are indispensable for using encoder-decoders in KBQA.

\paragraph{Joint Entity Linking}
Existing studies on KBQA typically rely on off-the-shelf entity linkers built upon simple techniques like fuzzy string matching to identify topic entities during pre-processing. We argue that the current norm is questionable because 1) pipelines suffer from error propagation and 2) it prohibits entity linking and semantic parsing from boosting each other. Preliminary efforts have been made to break the norm via relation-enhanced entity disambiguation~\cite{ye2021rng} and joint linking over relations and entities (but limited to a set of candidates filtered in advance). We envision a KBQA model that jointly performs entity linking and semantic parsing in a truly end-to-end manner. A promising direction is to project entities into a continuous space and perform differentiable operations over them~\cite{ren2021lego} because operating over millions of entities in discrete space can be intractable.



\subsection{Towards KBQA with Pre-Training} \label{subsec:plms}
\paragraph{KBQA with PLMs}
The high volume of KB schema items prohibits KBQA from jointly encoding the question and all schema items using PLMs via input concatenation, as done in text-to-SQL. To adapt it to KBQA, \citet{gu2020beyond} and \citet{chen-etal-2021-retrack} propose to narrow down the size of candidate schema items to fit PLMs' length limit. However, identifying relevant schema items beforehand is inflexible\nop{ and prone to error propagation}. A more promising direction is to integrate PLMs with constrained decoding~\cite{liang-etal-2017-neural, gu2022arcaneqa}, where relevant schema items are selected on the fly. In addition, to better understand the KB schema, future works may explicitly model the relationships among schema items and question tokens \cite{wang-etal-2020-rat}, instead of only delegating it implicitly to PLMs' self-attention layers via input concatenation. For encoder-decoder PLMs, \citet{xie2022unifiedskg} have shown that directly fine-tuning T5 suffices to outperform the prior art on \ComplexQ. Their preliminary studies indicate the great potential of using encoder-decoder PLMs as a unified solution to semantic parsing. Augmenting the encoder-decoder PLMs with constrained decoding algorithms like \Picard~\cite{scholak2021picard} is a promising direction to pursue in KBQA.

\nop{\citet{ye2021rng} uses T5 to refine incomplete queries. \citet{das-etal-2021-case} uses \BigBird~\cite{bigbird} to support long input comprising the question and retrieved cases. Impressively, \BigBird performs competitively even without retrieving any case\nop{, suggesting its great potential in KBQA}. \citet{xie2022unifiedskg} further demonstrate the capability of encoder-decoder PLMs using \Picard over T5. They outperform the prior art on \ComplexQ. However, these are still preliminary studies and the best practice for using PLMs remains an open question. }


\paragraph{KBQA-specific Pre-training}
Existing works in KBQA have shown the feasibility of cross-dataset pre-training~\cite{gu2020beyond,cao2021program}, while a general solution towards KBQA-specific pre-training like \Grappa for text-to-SQL remains absent. Applying task-specific pre-training in KBQA is challenging. First, synthesizing an aligned corpus of question-query pairs may lead to data contamination, especially for evaluating zero-shot generalization in KBQA. Second, the KBQA-specific pre-training task needs to take account of the structured information in queries, while binary classification tasks like SSP will fall short of this goal. A possible direction to address them is to synthesize KB-agnostic queries and specify a structured prediction task for pre-training.

\nop{we suggest synthesizing the queries using KB-agnostic descriptions for schema items (e.g., \textit{game series} instead of \textrelation{cvg.computer\_videogame.game\_series} in \Freebase) and directly using semantic parsing as the pre-training task if efficiency is not an issue.  }



\subsection{Towards More Generalizable KBQA}
\nop{Similar to text-to-SQL~\cite{yu-etal-2018-spider},} Similarly, research on KBQA has recently shifted its focus to non-i.i.d. generalization. \citet{gu2020beyond} release \Grail to systematically investigate the generalizability on three levels: \emph{i.i.d.}, \emph{compositional}, and \emph{zero-shot}. The key difference between non-i.i.d. generalization in text-to-SQL and KBQA is that the relevant tables are given as input in text-to-SQL, while all questions share the same KB in KBQA, so the model needs to determine which part of the KB is relevant by itself. This is extremely challenging for a non-i.i.d. setting because the model can easily overfit the KB segments seen during training.  Thus, \citet{gu2020beyond} suggest that highly generalizable models should feature effective search space pruning.  A promising direction is to use a more KB-specific constrained decoding algorithm for encoder-decoder PLMs rather than \Picard\cite{scholak2021picard, xie2022unifiedskg}\nop{, which achieves the best performance on the i.i.d. dataset \ComplexQ, while significantly underperforms \citet{ye2021rng} on \Grail}. 

Though the definitions in \citet{gu2020beyond} are extendable to KBQA with multiple KBs, their dataset is collected with a single KB (i.e., \Freebase). Recent works have addressed another type of non-i.i.d. generalization, i.e., evaluating on a KB different from training~\cite{ravishankar2021two,cao2021program}. \nop{For example, \citet{ravishankar2021two} first generate a KB-agnostic query and then ground the query to the target KB. }Developing KBQA models with such cross-KB generalizability is a stepping stone towards the ambitious goal of question answering over the linked open data (LOD) cloud~\cite{soru2020linked}. A dataset that supports systematic evaluation for such cross-KB generalization is in demand.


\subsection{Other Trends}
Several other trends in semantic parsing may also inspire research in KBQA. First, the human-in-the-loop methodology can effectively improve the accuracy of semantic parsing on complex questions that are challenging to solve in one shot~\cite{gur2018dialsql,yao-etal-2019-model}. Future works may study interactive KBQA to better handle complicated KB queries\cite{mo2021towards}. Second, prompting~\cite{gpt3} has been successfully applied to semantic parsing~\cite{schucher-etal-2022-power,yang2022seqzero}, while techniques like prompt tuning~\cite{lester-etal-2021-power} in KBQA remain to be investigated. Related to this point, recent works have considered a few-shot setting for cross-domain generalization in text-to-SQL~\cite{lee-etal-2021-kaggledbqa}, future works may also explore the potential of few-shot in-context learning and prompting in non-i.i.d. generalization in KBQA.

\nop{
\subsection{Other Directions}
Other promising directions on KBQA include but are not limited to 1) interactive KBQA aiming at complex queries via multi-turn conversations\nop{ instead of answering the question in a one-shot fashion}~\cite{mo2021towards}, 2) KBQA with incomplete KBs via differentiable operations~\cite{ren2021lego},\footnote{This is a particularly interesting direction which has recently received increasingly more attention. We will discuss it more in the extended version upon acceptance.} and---last but not least---3) standardization efforts, including building infrastructures (e.g., a shared platform for different communities working on KBQA)~\cite{perevalov2022knowledge} and benchmarking different supervisions or meaning representations.
}

\nop{PLM\\
Beam search based on encoder decoder\\
Task-specific Pre-training\\
Interaction \\
Joint entity linking and semantic parsing \\
Hybrid QA \\
differentiate reasoning (william cohen)\\
Linked open data}



\bibliography{sample}
\bibliographystyle{plainnat}

\end{document}